\documentclass[pdflatex,bst/sn-mathphys-num]{sn-jnl}%
\usepackage{graphicx}%
\usepackage{multirow}%
\usepackage{amsmath,amssymb,amsfonts}%
\usepackage{amsthm}%
\usepackage{mathrsfs}%
\usepackage[title]{appendix}%
\usepackage{xcolor}%
\usepackage{textcomp}%
\usepackage{manyfoot}%
\usepackage{booktabs}%
\usepackage{algorithm}%
\usepackage{algorithmicx}%
\usepackage{algpseudocode}%
\usepackage{listings}%
\usepackage{soul} %

\usepackage{subcaption}

\usepackage{natbib} 
\theoremstyle{thmstyleone}%

\theoremstyle{definition}

\theoremstyle{thmstyletwo}%

\theoremstyle{thmstylethree}%

\begin{document}

\title[Article Title]{Classification of Epileptic iEEG using Topological Machine Learning}

\author[1, 4]{\fnm{Sunia} \sur{Tanweer}}

\author[2, 3]{\fnm{Narayan Puthanmadam} \sur{Subramaniyam}}

\author*[4]{\fnm{Firas A.} \sur{Khasawneh}}\email{khasawn3@msu.edu}

\affil[1]{\orgdiv{Department of Mechanical Engineering}, \orgname{Michigan State University}, \city{East Lansing}, \postcode{48823}, \state{MI}, \country{USA}}

\affil[2]{\orgdiv{Faculty of Medicine and Health Technology}, \orgname{Tampere University}, \city{Tampere}, \country{Finland}}

\affil[3]{\orgdiv{A.I.Virtanen Insttute of Molecular Sciences}, \orgname{University of Eastern Finland}, \city{Kuopio}, \country{Finland}}

\affil[4]{\orgdiv{Department of Computational Mathematics, Science and Engineering}, \orgname{Michigan State University}, \orgaddress{\city{East Lansing}, \postcode{48823}, \state{MI}, \country{USA}}}

\abstract{
Epileptic seizure detection from EEG signals remains challenging due to the high dimensionality and nonlinear, potentially stochastic, dynamics of neural activity. In this work, we investigate whether features derived from topological data analysis (TDA) can improve the classification of brain states in preictal, ictal and interictal iEEG recordings from epilepsy patients using multichannel data. We analyze data from 55 patients, significantly larger than many previous studies that rely on patient-specific models. Persistence diagrams derived from iEEG signals are vectorized using several TDA representations, including Carlsson coordinates, persistence images, and template functions. 
To understand how topological representations interact with modern machine learning pipelines, we conduct a large-scale ablation study across multiple iEEG frequency bands, dimensionality reduction techniques, feature representations, and classifier architectures.
Our experiments show that dimension-reduced topological representations achieve up to 80\% balanced accuracy for three-class classification. Interestingly, classical machine learning models perform comparably to deep learning models, achieving up to 79.17\% balanced accuracy, suggesting that carefully designed topological features can substantially reduce model complexity requirements. In contrast, pipelines preserving the full multichannel feature structure exhibit severe overfitting due to the high-dimensional feature space. These findings highlight the importance of structure-preserving dimensionality reduction when applying topology-based representations to multichannel neural data.
}

\keywords{epilepsy, classification, seizure, topological data analysis, machine learning}

\maketitle

\section{Introduction}

Epilepsy is a chronic neurological disorder characterized by recurrent seizures, which are sudden and unpredictable disruptions of normal brain activity. Electroencephalography (EEG) remains the primary tool for diagnosing and monitoring epileptic patients due to its ability to directly record brain activity with millisecond temporal resolution. 

Traditionally, neurologists identify epileptic seizures based on specific patterns in the EEG waveforms, including spikes, sharp waves, and rhythmic activity. However, manual analysis of EEG recordings is time-consuming, subjective, and increasingly impractical given the increasing availability of long-duration EEG recordings. This has created a need for automated methods that can assist clinicians in detecting and characterizing seizure-related activity. 

Distinguishing between interictal (between seizures), preictal (before a seizure), and ictal activity (seizure) is important for developing an automated classification system that will considerably reduce the burden of manual annotation~\cite{Roy2021LancetAI}. Furthermore, identification of preictal state is crucial for designing seizure prediction systems, that will enable timely intervention. Although scalp EEG is non-invasive and widely used, it is limited by low signal-to-noise ratio (SNR) \cite{Tito2009} and has poor spatial resolution. Intracranial EEG (iEEG) recordings, obtained by placing electrode grid directly on the cortex (electrocorticography - ECoG) or by using depth electrodes (stereo-EEG), has higher SNR and spatial specificity, making it well suited for epileptic EEG analysis.

In recent years, machine learning approaches have increasingly been employed to assist in the automatic classification of epileptic states~\cite{Tuncer2022, Ryu2021, panuccio2024using}. These methods aim to improve diagnostic accuracy and efficiency by learning from labelled EEG data, distinguishing between ictal , preictal  and interictal states. Machine learning and, more recently, deep learning have become a cornerstone of modern EEG analysis~\cite{Yogarajan2023, Chandel2023, Craley2021, Ansari2019, Li2023, Shen2024, Li2020, Liu2019, Chen2023}. Traditional methods for EEG classification include support vector machines (SVMs), random forests, and neural networks~\cite{Yogarajan2023, Chandel2023, Wang2023,Wang2023a}, which are trained on features such as frequency bands~\cite{Wang2022}, wavelet coefficients~\cite{Akut2019}, and statistical measures~\cite{Rashid2017, Alalayah2023} of EEG data. However, capturing the non-linear and high-dimensional dynamics of EEG signals remains challenging, particularly in the case of multi-channel recordings. 

Topological Data Analysis (TDA) has recently emerged as a promising framework for analyzing complex signals such as EEG recordings. TDA provides tools for extracting structural information from data by analyzing the shape of signals across multiple scales. Persistent homology, one of the central techniques in TDA, summarizes the evolution of topological features through persistence diagrams, which record the birth and death of features as a filtration parameter varies. These diagrams can then be converted into vector representations suitable for machine learning models. The combination of TDA and machine learning has been particularly successful in single-channel EEG datasets. TDA-based features, such as barcodes and persistence diagrams, have been introduced as a new form of input for classifiers. These features capture the shape and topology of the EEG signal, offering an alternative to conventional feature extraction methods. Early work applying TDA to EEG data primarily focused on single-channel recordings. In these cases, topological features derived from persistence diagrams were used as inputs to machine learning classifiers, achieving promising results in terms of accuracy~\cite{Tuncer2022, Ryu2021}. Many non-TDA features have also been used and have resulted in high accuracies. Some of these features are Permutation Entropy, Hjorth parameters, Fractal Dimension, Approximate Entropy, Hurst Exponent, Correlation Dimension, Lyapunov Exponents, Zero Crossing Rate etc.~\cite{Natu2022, Yacoob1996, Winterhalder2003}. 

However, the transition from single-channel to multi-channel EEG classification introduces additional complexity, as multiple interacting signals need to be analyzed simultaneously. Multi-channel EEG data provides a richer source of information but also presents significant challenges due to the increase in dimensionality and the complexity of interactions between channels. Combining data from multiple channels for classification purposes requires advanced techniques of feature fusion, data fusion or dimensionality reduction. Traditional methods struggle to efficiently aggregate information across channels without losing important temporal and spatial dependencies. Even then, several studies have demonstrated the effectiveness of TDA and conventional signal processing in capturing the dynamics of epileptic seizures. For instance, Wang et al.~\cite{Wang2023} used persistent homology with GoogleNet (a deep-learning framework) to detect seizure transitions and found that TDA-based features provide robust representations of EEG signals that can be used for classification. Recent work by Hu et al.~\cite{Hu2024} reached a very high accuracy of 91\% with an iterative gated graph convolution network. However, what is common among all the available multi-channel methods is the use of deep learning for classification. Although deep learning can result in high accuracy, it is a highly time-consuming approach, requiring large datasets and substantial computational resources for training. 

In this work, we investigate whether topological features derived from iEEG signals can provide effective representations for seizure state classification across a relatively large cohort of patients. We analyze iEEG recordings from 55 epilepsy patients and classify signal segments into interictal, preictal, and ictal states using machine learning pipelines built on topological features. Several vectorization techniques for persistence diagrams are considered, including Carlsson coordinates~\cite{carlsson2009}, persistence images~\cite{adams2017}, and template functions~\cite{Tymochko2019}.

To better understand how topological features interact with modern machine learning pipelines, we conduct a comprehensive ablation study across multiple components of the classification pipeline. These components include iEEG frequency bands, dimensionality reduction methods, topological feature representations, and classifier architectures. Both classical machine learning models and deep learning models are evaluated in order to assess the extent to which classifier complexity influences performance when using topology-based features. The results demonstrate that topology-based features combined with dimensionality reduction can achieve strong classification performance across multiple seizure states. In addition, the study highlights the importance of dimensionality reduction when applying topology-based representations to multichannel neural data, as extremely high-dimensional feature spaces can lead to severe overfitting. These findings provide insights into how topological representations can be effectively integrated into machine learning pipelines for iEEG analysis. 

\section{Mathematical Background}\label{sec:eeg_maths}

\subsection{Sublevel Persistence on 1D signals}

\begin{figure}[!htbp]
\centering
\includegraphics[width=0.8\linewidth]{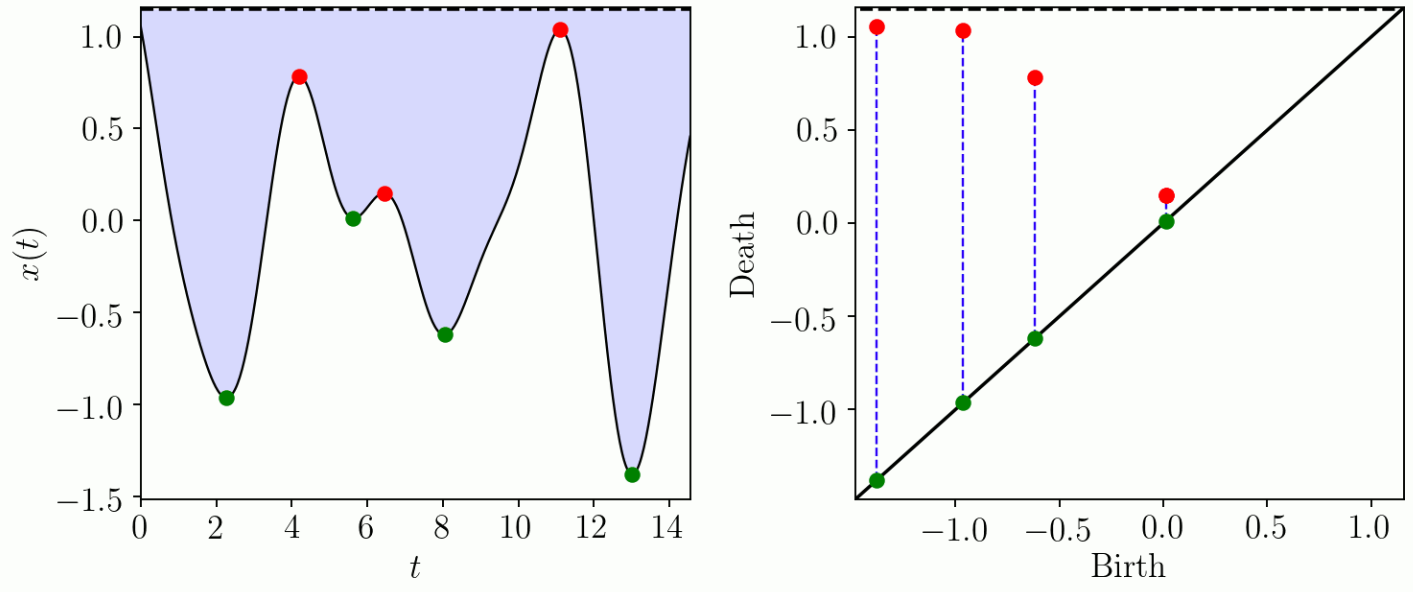}
\caption{Sublevel Persistence on 1D signals.}
\label{fig:sublevel_signals}
\end{figure}

Given a function $f : [0, T] \xrightarrow{} \mathbb{R}$ and a real number $r \in \mathbb{R}$, the sublevel set for $r$ is defined as $f^{-1}(-\infty,r]$. As the parameter $r$ (also called filtration paramater) increases, the sublevel sets may grow but remain the same until we reach a local extrema. If the extrema reached is a local minimum, a new set is born at $r_B$. If the extrema is a local maximum, two previously existing sets are combined into one. If the two sets were $r_B$ and $r'_{B}$, such that $r_B \leq r'_{B}$, with the maximum reached at $r_D$, it is said that the component born at $r'_{B}$ dies going into $r_D$. Hence, the pair $(r'_{B}, r_D)$ is called a persistence pair, with $r'_{B}$ being the location of birth and $r_{D}$ the location of death. As $r$ is varied between $-\infty$ and $\infty$, the persistence diagram is the collection of all $n$ such pairs, $\mathbb{D}(f) = \{(b_i, d_i)\}^n_{i=1}$. For each of these pairs, the value $d_i - b_i$ is termed the life of that component. See Fig.~\ref{fig:sublevel_signals} for an example.

In the context of EEG signals, persistent homology provides a way to capture structural patterns in the signal beyond traditional spectral or statistical descriptors, allowing the identification of stable geometric features in the temporal dynamics.

\subsection{Featurization with Persistence Diagrams}

A key challenge in TDA is converting topological information---usually in the form of persistence diagrams---into a format that can be effectively used by machine learning algorithms. Several methods have been proposed for vectorizing persistence diagrams while preserving their essential topological features. Below, we outline some of the prominent techniques: Carlsson coordinates, persistence images, and template functions (tent and polynomial functions). These vectorization techniques allow persistence diagrams to be incorporated into standard machine learning pipelines while preserving the stability properties of persistent homology representations.

\subsubsection{Carlsson Coordinates}

Carlsson coordinates~\cite{carlsson2009} are a set of five functions $f: \mathbb{R}^2 \xrightarrow{} \mathbb{R}$ defined for a persistence diagram. For a persistence diagram with $n$ points $(b_i,d_i)$, the functions are defined as follows:
\begin{equation}
    f_1 = \sum_{i = 1}^n b_i(d_i - b_i)
\end{equation}
\begin{equation}
    f_2 = \sum_{i = 1}^n (d_{\text{max}} - d_i)(d_i - b_i)
\end{equation}
\begin{equation}
    f_3 = \sum_{i = 1}^n b^2_i(d_i - b_i)^4
\end{equation}
\begin{equation}
    f_4 = \sum_{i = 1}^n (d_{\text{max}} - d_i)^2(d_i - b_i)^4
\end{equation}
\begin{equation}
    f_5 = \text{max}(d_i - b_i)
\end{equation}

\subsubsection{Persistence Images}

Persistence images~\cite{adams2017} are a widely used method for transforming persistence diagrams into feature vectors that are compatible with machine learning models. This approach involves assigning a Gaussian distribution to each persistence point, creating a smooth image from the persistence diagram. The image is then discretized into a fixed-size grid, enabling direct input into algorithms such as support vector machines or deep neural networks. Persistence images are particularly advantageous because they provide a stable and continuous representation of topological features, which can be effectively used in EEG classification tasks to capture neural dynamics over time.

Mathematically, the persistence diagram $\mathcal{D} = \{(b_i, d_i)\}_{i=1}^{n}$ is mapped into birth-death coordinates, and each point $(b_i, d_i)$ contributes to the image by placing a Gaussian distribution centered at $(b_i, d_i)$. The weighted sum of these Gaussians forms the persistence image, which can be adjusted through parameter tuning of the Gaussian kernel.

\subsubsection{Template Functions (Tent and Polynomial Functions)}

Template functions~\cite{Tymochko2019} offer a flexible way to vectorize persistence diagrams by applying functions that map birth-death pairs into scalar values. Two common template functions are tent functions and polynomial functions, each offering a different way to emphasize certain parts of the persistence diagram.

\begin{itemize}
    \item \textbf{Tent Functions}: These are triangular-shaped functions defined over the plane of the persistence diagram. For each birth-death pair $(b_i, d_i)$, the function takes the value of the persistence (i.e., $d_i - b_i$) within a certain region and decreases linearly towards the boundary of that region. Tent functions emphasize localized information in the diagrams.
    \item \textbf{Polynomial Functions}: Polynomial functions map each persistence point $(b_i, d_i)$ to a polynomial of their birth and death times. Higher-order polynomial functions allow the user to emphasize global information in the diagrams.
\end{itemize}

Template functions offer a computationally simple and interpretable way to vectorize topological features, making them effective for EEG classification tasks, where capturing specific birth and death events can reflect brain activity transitions.

\section{Methods}

\begin{figure}[!htbp]
\centering
\includegraphics[width=\linewidth]{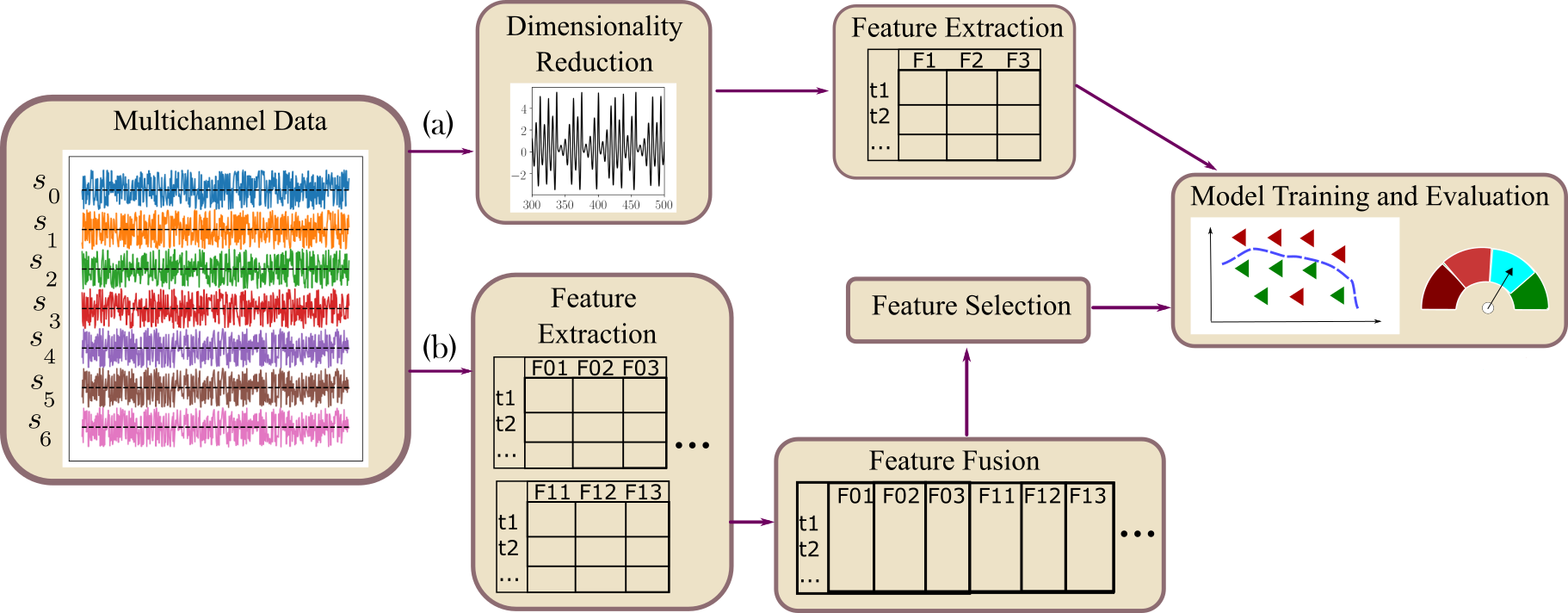}
\caption{Overview of the two classification pipelines considered in this study: (a) Dimension-reduced and (b) Multichannel preserving channel-wise features.}
\label{fig:pipeline_overview}
\end{figure}

In this work, we evaluated two ML-classification pipelines from multichannel iEEG recordings. Figure~\ref{fig:pipeline_overview}a shows the first pipeline which applies dimensionality reduction prior to classification and henceforth we will refer to it as the \emph{dimension-reduced pipeline}. The second pipeline shown in Fig.~\ref{fig:pipeline_overview}b preserves channel-wise topological structure as long as possible and we will refer to it as the \emph{multichannel pipeline}. 

\subsection{Dataset and classification task}

The experiments were conducted on multichannel iEEG recordings from 55 epilepsy patients, taken from HUP dataset~\cite{HUP_dataset}. The dataset has samples assigned to one of three classes: interictal, preictal, or ictal. The goal of the classification task was therefore a three-class prediction problem based on topological representations of the iEEG signals.

The iEEG recordings contain multiple channels representing electrical activity measured across different regions of the brain. Since seizure-related information may be distributed across channels and frequency ranges, both reduced and channel-preserving representations were considered.

\subsection{Preprocessing}

Prior to feature extraction, iEEG recordings were filtered into commonly used frequency bands in order to isolate physiologically meaningful components. The following bands were considered: Delta (0.5--4 Hz), Theta (4--8 Hz), Alpha (8--13 Hz), Beta (13--30 Hz), Low Gamma (30--50 Hz), High Gamma (50--100 Hz). In addition to the filtered signals, the raw broadband iEEG signal was also evaluated. Each filtered version of the signal was passed independently through the subsequent feature extraction and classification stages.

\subsection{Feature Extraction}

For each iEEG segment, persistence diagrams were computed using sublevel set persistence on one-dimensional signal representations. These persistence diagrams encode the birth and death of topological features in the signal as the filtration parameter varies. To incorporate persistence diagrams into machine learning models, the diagrams were vectorized using several topological feature representations. The feature sets considered in this work were Carlsson coordinates, Persistence images, Template functions (Tent) and Template Functions (Polynomial). For baseline comparison, entropy-based features of complexity entropy, Renyi entropy, Tsallis entropy, and weighted permutation entropy were used.

\subsection{Pipeline I: Dimension-Reduced Classification}

In the dimension-reduced pipeline, the multichannel iEEG signal was transformed into a one-dimensional representation before featurization. After filtering into a specified frequency band, dimensionality reduction was applied, followed by extraction of features and supervised classification.

A broad range of dimensionality reduction methods was evaluated, including linear and nonlinear. The tested methods included principal component analysis (PCA), linear discriminant analysis (LDA), non-negative matrix factorization (NMF), factor analysis (FA), truncated singular value decomposition (TSVD), Isomap, locally linear embedding (LLE), multidimensional scaling (MDS), and t-SNE.

The resulting lower-dimensional representations were then featurized and passed to a classifier.

\subsection{Pipeline II: Multichannel Classification}

In the multichannel pipeline, channel-wise information was preserved by computing features separately for each iEEG channel. The resulting features were then concatenated into a single multichannel feature vector prior to classification. This strategy retains more data and structural information than the dimension-reduced pipeline, but it also leads to a much larger feature space. Depending on the chosen feature representation, the dimensionality of the resulting multichannel feature vector ranged from several hundred to several thousand features. 

Because the number and identity of iEEG channels varied across recordings in the dataset, it was necessary to ensure a consistent channel set across all samples used in the multichannel pipeline. To achieve this, we selected a subset of recordings sharing a common set of channels while maximizing both the number of available samples and the number of retained channels. As a result, the multichannel experiments were conducted using a slightly reduced subset of the full dataset in order to maintain consistent channel alignment across all samples.
Then, to mitigate the resulting curse of dimensionality, feature selection and strong regularization were applied. In particular, SelectKBest was used to retain at most 500 features prior to classification. Even with this reduction, the multichannel feature space remained high-dimensional relative to the number of available training samples.

\subsection{Machine Learning Models}

Both classical machine learning models and deep learning models were evaluated in order to assess the role of classifier complexity when combined with topological features. The classical machine learning models included Logistic Regression, Ridge Classifier, Linear SVC, SVC with RBF kernel, Random Forest, Gradient Boosting, Gaussian Naive Bayes, Linear Discriminant Analysis, and Multi-layer Perceptron. The deep learning models included several fully connected and convolution-based architectures, including DeepMLP variants, ResNet-based models, attention-based multilayer perceptrons, and Conv1DMLP models. For the multichannel pipeline, regularized deep learning architectures were considered in order to reduce overfitting. Details are found in Appendix~\ref{sec:appendix}.

Model performance was evaluated using balanced accuracy as the primary metric. For both pipelines, the dataset was divided into training and test sets using an 80/20 split. Performance on the held-out test set was used for final evaluation. In addition to test performance, the gap between training and test balanced accuracy was monitored in order to assess overfitting; any model with more than 10\% difference was dropped from the summary results.

\section{Results}

\subsection{Overall Performance}

Two primary modeling strategies were evaluated: a dimension-reduced pipeline and a multichannel pipeline preserving spatial channel structure. The dimension-reduced pipeline achieved substantially higher performance, with a maximum balanced accuracy of 80\%. Table~\ref{tab:best_conf} summarizes the top 10 performing configurations in the dimension-reduced pipeline. In contrast to dimension-reduced pipeline, the multichannel pipeline achieved a maximum balanced accuracy of 59.5\%. These results indicate that high-dimensional multichannel feature spaces introduce significant challenges for model generalization. 

\begin{table}[htbp]
\centering
\caption{Top 10 performing configurations in the dimension-reduced pipeline.}
\begin{tabular}{cccccc}
\hline
Rank & Accuracy & Band & Dim Red & Feature & Classifier \\
\hline
1 & 80.00\% & Low Gamma & FA & Carlsson & Conv1DMLP \\
2 & 79.17\% & Beta & FA & Carlsson & Logistic Regression \\
3 & 77.78\% & Delta & LLE & Carlsson & Logistic Regression \\
4 & 77.78\% & Delta & LLE & Carlsson & LinearSVC \\
5 & 77.08\% & Beta & FA & Carlsson & LinearSVC \\
6 & 73.67\% & Alpha & NMF & Polynomial Template & MLP (sklearn) \\
7 & 73.33\% & Low Gamma & FA & Tent Template & MLP (sklearn) \\
8 & 72.92\% & Beta & FA & Carlsson & LDA \\
9 & 72.92\% & Beta & FA & Polynomial Template & DeepMLP (Large) \\
10 & 72.12\% & Delta & t-SNE & Tent Template & Conv1DMLP \\
\hline
\end{tabular}
\label{tab:best_conf}
\end{table}

\subsection{Deep Learning vs Classical Machine Learning}

\begin{figure}[!htbp]
\centering
\includegraphics[width=0.95\linewidth]{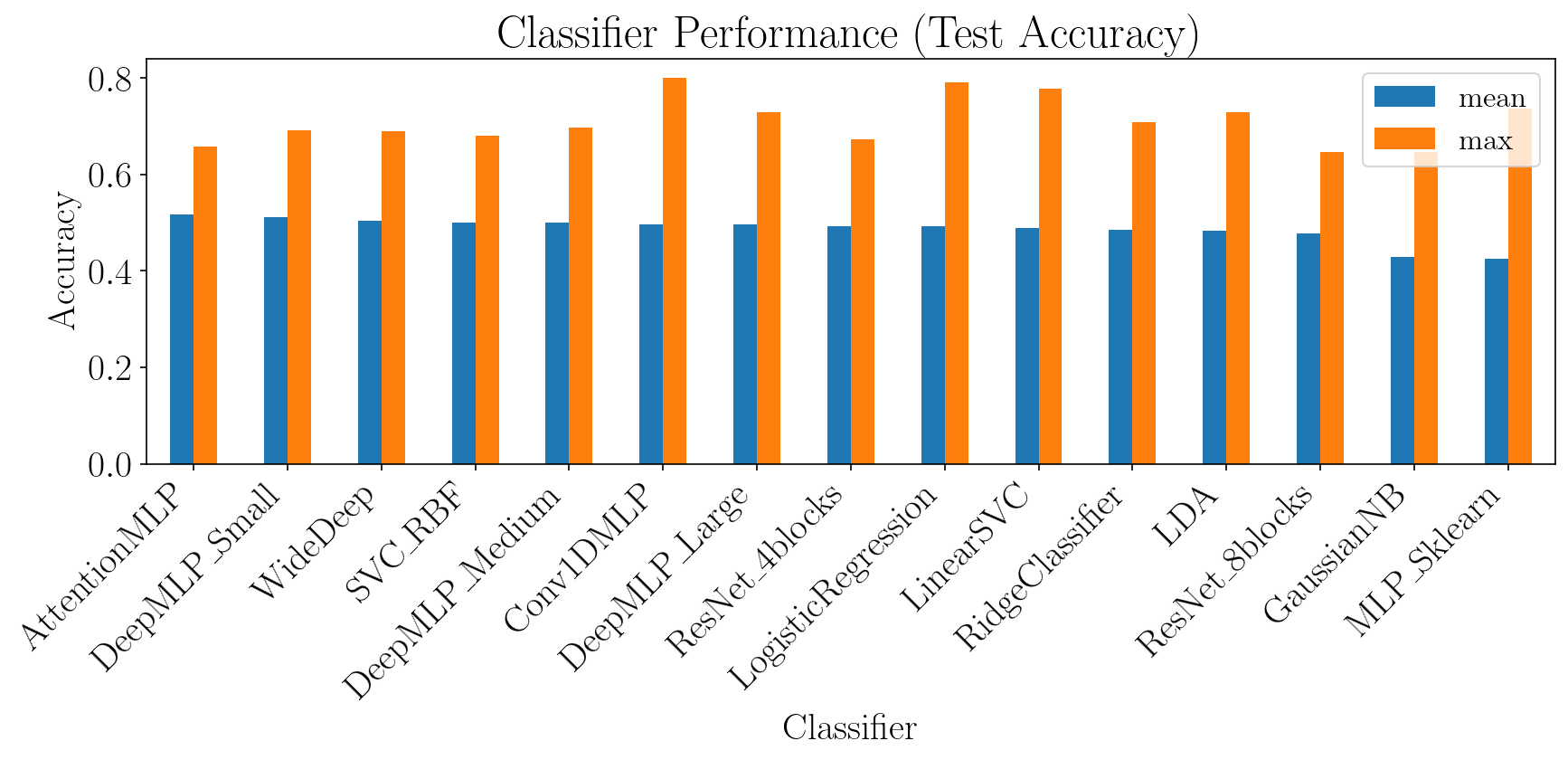}
\caption{Comparison of balanced accuracy between classical machine learning models and deep learning models across all configurations.}
\label{fig:model_comparison}
\end{figure}

Both classical machine learning models and deep learning architectures were evaluated. The best-performing deep learning configuration achieved 80\% balanced accuracy using low gamma band signals and factor analysis for dimensionality reduction, with Carlsson coordinate features, and a Conv1D MLP classifier. Interestingly, classical machine learning models achieved comparable performance. The best classical model, logistic regression combined with Carlsson coordinates and factor analysis, achieved 79.17\% balanced accuracy. These results suggest that carefully constructed topological feature representations can enable simple classifiers to perform competitively with more complex neural network architectures.
Figure~\ref{fig:model_comparison} compares the performance of deep learning models and classical machine learning models across all configurations. While deep learning models achieved slightly higher maximum performance, the overall performance distributions of the two model classes were similar. This result suggests that well-constructed topological feature representations can allow relatively simple models to achieve performance comparable to more complex neural network architectures. This observation is particularly relevant for practical clinical pipelines, where simpler models may offer lower computational cost, easier tuning, and greater interpretability without substantial loss in predictive performance.

\subsection{Frequency Band Analysis}

Performance varied across frequency bands. The delta band achieved the highest mean balanced accuracy (50.20\%), indicating that low-frequency oscillations contain useful information for seizure classification. However, the best single-performing configuration was obtained using the low gamma band (30-50 Hz), suggesting that higher frequency activity may capture seizure-related dynamics under certain model configurations.

\begin{figure}[!htbp]
\centering
\includegraphics[width=0.8\linewidth]{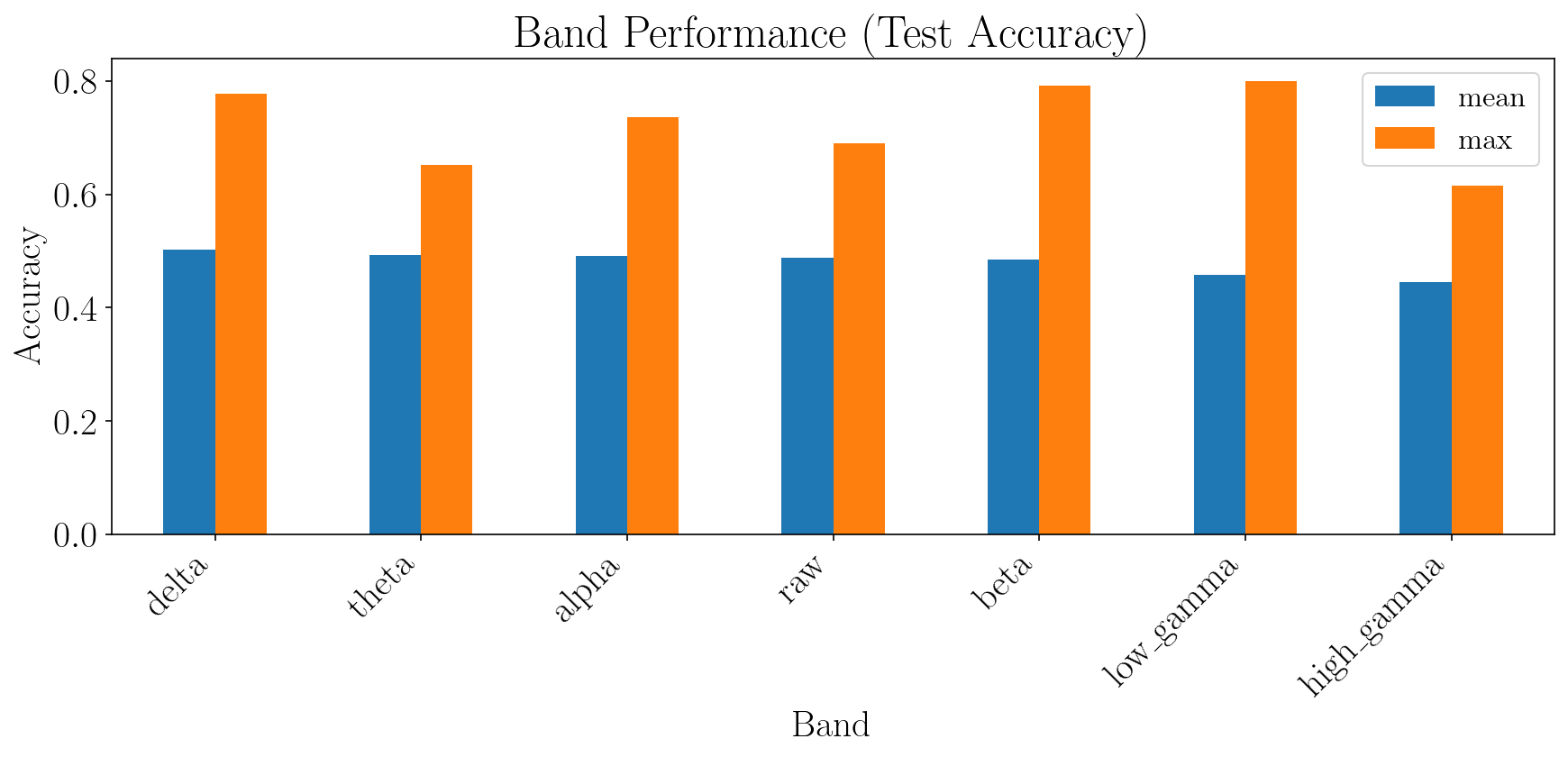}
\caption{Comparison of balanced accuracy between frequency bands across all configurations.}
\label{fig:band_comparison}
\end{figure}

The relationship between frequency band and classification performance is illustrated in Fig.~\ref{fig:band_comparison}.

\subsection{Dimensionality Reduction Methods}

Dimensionality reduction played a critical role in classification performance. Nonlinear manifold learning techniques such as t-SNE achieved the highest mean performance across configurations, while Factor Analysis produced the best single-performing pipeline when combined with deep learning models. Linear methods such as PCA and LDA also provided stable performance across configurations, indicating that relatively simple dimensionality reduction methods can effectively compress topological feature representations.

\begin{figure}[!htbp]
\centering
\includegraphics[width=0.8\linewidth]{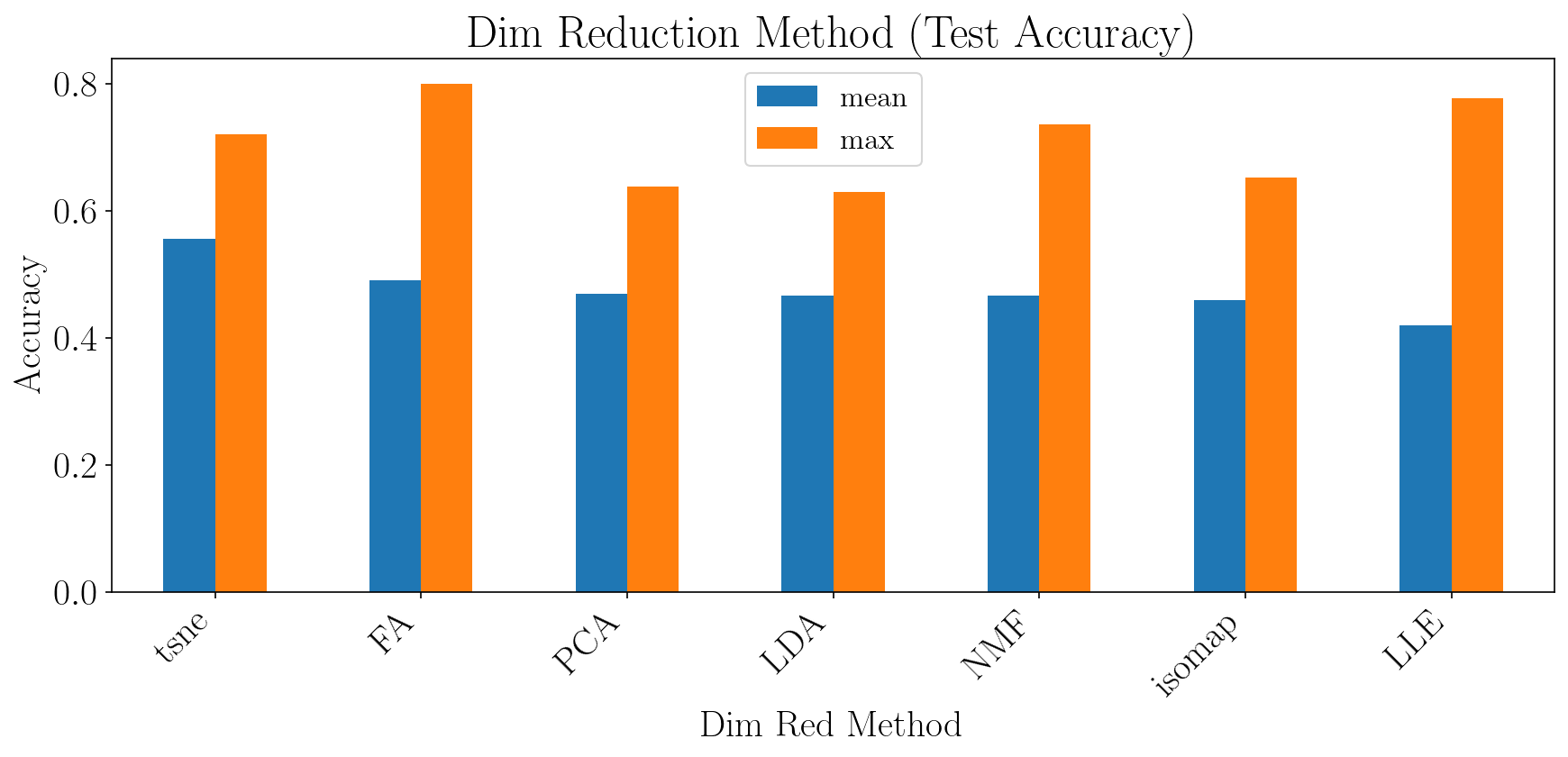}
\caption{Comparison of balanced accuracy between dimensionality reduction techniques across all configurations.}
\label{fig:dimred_comparison}
\end{figure}

Figure~\ref{fig:dimred_comparison} compares the maximum balanced accuracy obtained using different dimensionality reduction techniques.

\subsection{Topological Feature Comparison}

Among the evaluated topological features, polynomial template functions achieved the highest mean performance, while Carlsson coordinates achieved the best individual result. Overall, topology-based features consistently outperformed the entropy-based baseline features, indicating that persistent homology captures structural properties of iEEG signals that are relevant for seizure classification.

\begin{figure}[!htbp]
\centering
\includegraphics[width=0.8\linewidth]{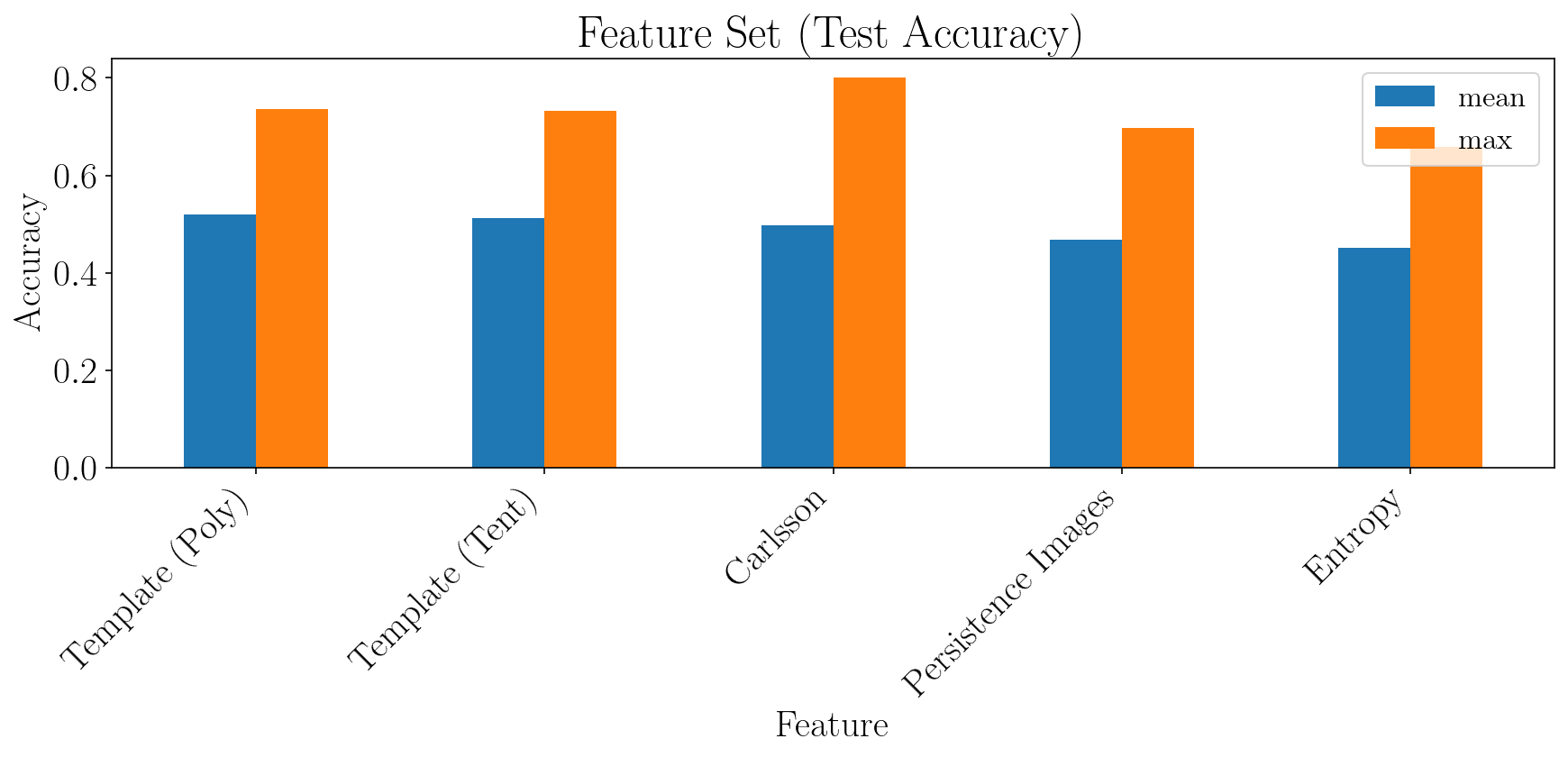}
\caption{Comparison of balanced accuracy between various feature sets across all configurations.}
\label{fig:feature_comparison}
\end{figure}

\subsection{Multichannel Pipeline Performance}

The multichannel pipeline preserved the spatial structure of iEEG channels by computing topological features for each channel independently. However, this approach resulted in extremely high-dimensional feature spaces.
Even after aggressive dimensionality reduction and regularization, models trained on the multichannel feature representation exhibited severe overfitting. The best-performing model, without overfitting, achieved 59.5\% balanced accuracy using a regularized deep multilayer perceptron classifier.
These results suggest that the number of samples available in the dataset is insufficient for reliable learning in such high-dimensional spaces---highlighting the curse of dimensionality in high-dimensional data.

\section{Discussion}

The results demonstrate that topological features extracted from iEEG signals provide effective representations for seizure state classification, particularly when combined with dimensionality reduction. While prior work has established the promise of topological data analysis (TDA) for EEG classification, our findings clarify how these representations behave in multichannel, multi-patient settings.

Previous studies have reported strong performance of TDA-based features for epileptic EEG analysis, particularly in single-channel or patient-specific settings. For example, Tuncer et al.~\cite{Tuncer2022} and Ryu et al.~\cite{Ryu2021} demonstrated that persistence-based features can successfully distinguish seizure states when applied to individual channels. More recent approaches have incorporated persistent homology into deep learning pipelines or graph-based neural networks to handle multichannel data, often achieving high accuracy on benchmark datasets \cite{Wang2023, Hu2024}. However, these studies typically rely on relatively small datasets, patient-specific models, or controlled benchmark settings, which may limit their generalizability. In contrast, this work considers a cohort of 55 patients in a three-class classification setting, making the problem substantially more challenging and clinically relevant. The achieved performance of up to 80\% balanced accuracy is therefore competitive in a patient-independent, multichannel context.

A key contribution of this work is the explicit identification of feature dimensionality as a primary bottleneck in multichannel TDA pipelines. While prior work has demonstrated the effectiveness of persistent homology representations, relatively little attention has been paid to how these features scale with increasing numbers of channels. The results show that naive concatenation of channel-wise topological features leads to extremely high-dimensional feature spaces, resulting in severe overfitting even with regularization. This observation provides a possible explanation for the widespread reliance on deep learning in multichannel EEG studies: deep architectures implicitly perform feature compression through hierarchical representations. In contrast, the present results demonstrate that explicit dimensionality reduction techniques, such as factor analysis and manifold learning, can effectively compress topological features while preserving discriminative structure. This suggests that limitations observed in prior multichannel TDA approaches may stem less from the representational capacity of topology and more from challenges associated with feature scaling.

Another important finding is that classical machine learning models perform comparably to deep learning architectures when trained on appropriately reduced topological features. This contrasts with much of the recent EEG literature, where deep learning is often assumed to be necessary for high performance \cite{Wang2025AIReview}. The results suggest that the quality of the feature representation plays a more critical role than classifier complexity. Topological features encode structural properties of the signal that are invariant to certain transformations and robust to noise, which may reduce the need for highly expressive models. From a practical perspective, this has important implications for clinical deployment, where simpler models offer advantages in interpretability, computational efficiency, and reliability. From a signal analysis perspective, the results supp ort the view that persistent homology captures structural properties of EEG signals beyond conventional spectral or entropy-based features. This is consistent with earlier work using persistence landscapes and related summaries to characterize seizure dynamics~\cite{Wang2018PL}, but extends these ideas to a multichannel, machine learning setting.

Despite these promising results, several limitations remain. The multichannel feature space remains high-dimensional relative to the available data, and spatial dependencies between channels are not explicitly modeled. Future work may address these issues through graph-based or simplicial representations of inter-channel relationships, as well as extensions toward seizure prediction using preictal dynamics. Overall, this study demonstrates that TDA-based features provide a robust and interpretable framework for EEG classification, while emphasizing the critical role of dimensionality reduction in enabling their effective use in multichannel settings.

\section{Conclusion}

We investigated the use of topological data analysis features for classification of seizure states in multichannel iEEG recordings from 55 epilepsy patients. A comprehensive ablation study was conducted across frequency bands, dimensionality reduction techniques, topological feature representations, and classifier architectures.
The results demonstrate that topology-based features combined with dimensionality reduction can achieve balanced accuracies of up to 80\% in three-class seizure classification. Classical machine learning models were found to perform competitively with deep learning architectures when trained on appropriately reduced feature representations.

In contrast, pipelines preserving the full multichannel feature structure exhibited severe overfitting due to the extremely high dimensionality of the resulting feature space---even with feature selection and heavy regularization. These findings highlight the importance of dimensionality reduction when applying topological representations to high-dimensional neural data.
Overall, this study provides evidence that topological representations offer a promising framework for analyzing iEEG signals and can form the basis for robust machine learning pipelines for seizure detection and classification.

\appendix
\section{Custom Model Architectures}
\label{sec:appendix}

This appendix describes the custom neural network architectures and training procedures used in both the single-channel and multichannel iEEG experiments. All models take a feature vector $x \in \mathbb{R}^d$ and output class logits $f(x) \in \mathbb{R}^C$.

\subsection{Deep Multilayer Perceptrons}

Several multilayer perceptron (MLP) architectures were evaluated for classification of the extracted feature representations. Each hidden layer applies a linear transformation followed by batch normalization, a ReLU activation, and dropout:

\[
h_{l+1} = \text{Dropout}(\text{ReLU}(\text{BatchNorm}(W_l h_l + b_l))).
\]

Three configurations were tested in the single-channel experiments: (1) Small ($128 \rightarrow 64$), (2) Medium ($256 \rightarrow 128 \rightarrow 64$), and (3) Large ($512 \rightarrow 256 \rightarrow 128 \rightarrow 64$). For the multichannel experiments, a strongly regularized variant was used with hidden dimensions $256 \rightarrow 128 \rightarrow 64$ and dropout rate 0.5.

\subsection{Residual Network}

To allow deeper nonlinear transformations while maintaining stable training, a fully connected residual network was implemented. The input features are first projected to a hidden representation:

\[
h_0 = \text{ReLU}(\text{BatchNorm}(W_0 x + b_0)).
\]

Each residual block computes

\[
h_{l+1} = \text{ReLU}(h_l + F(h_l)),
\]

where $F(\cdot)$ consists of two fully connected layers with batch normalization and dropout. The single-channel experiments evaluated networks with four and eight residual blocks. The multichannel experiments used a regularized variant with four residual blocks and dropout applied within each block.

\subsection{Attention-Based MLP}

To model interactions between feature dimensions, an attention-based architecture was implemented. The input features are first projected into a latent representation ($d_h=256$). A multi-head self-attention layer with eight heads is then applied, followed by a feed-forward network with dimensions $256 \rightarrow 512 \rightarrow 256$. Residual connections and layer normalization are used around both the attention module and the feed-forward network. The resulting representation is passed to a linear output layer.

\subsection{Wide \& Deep Network}

A Wide \& Deep architecture was used to capture both linear and nonlinear feature relationships. The wide component is a linear model

\[
f_{\text{wide}}(x) = W_w x + b_w,
\]

while the deep component is an MLP with hidden dimensions

\[
256 \rightarrow 128 \rightarrow 64.
\]

The final prediction is obtained by summing the outputs of the wide and deep components.

\subsection{1D Convolutional Feature Network}

To capture local dependencies between neighboring features, the input vector is reshaped into a one-dimensional sequence and processed by a small convolutional network. Two convolutional layers (32 and 64 channels with kernel size 3) are applied, each followed by batch normalization and ReLU activation. Global average pooling aggregates the feature maps, after which a fully connected layer with 128 hidden units produces the final logits.

\subsection{Feature Selection for High-Dimensional Representations}

For multichannel feature representations, dimensionality reduction was optionally applied after feature standardization to mitigate overfitting. The feature selection methods SelectKBest and PCA were used; no reduction for $d < 1000$, SelectKBest for moderate dimensionality, and PCA for very high-dimensional representations. Early stopping with a patience of 10-15 epochs was also applied to prevent overfitting.

\paragraph{Acknowledgements}{This material is based upon work supported by the Air Force Office of Scientific Research under award number FA9550-22-1-0007, the Army Research Lab under award number W911NF-26-1-A003 and NSF funded-Frontera Computational Science Fellowship.}

\paragraph{CRediT}{ST---Conceptualization, Data Curation, Formal Analysis, Funding Acquisition, Investigation, Methodology, Project Administration, Software, Validation, Visualization, Writing - Original Draft, Writing - Review \& Editing.
NPS---Conceptualization, Data Curation, Software, Writing - Review \& Editing.
FAK---Conceptualization, Funding Acquisition, Project Administration, Resources, Supervision, Writing - Review \& Editing.}

\paragraph{Data and Supplementary Materials}{The dataset used can be found here~\cite{HUP_dataset}: https://doi.org/10.18112/openneuro.ds004100.v1.1.3}

\bibliographystyle{unsrt}
\bibliography{bibliography}

\end{document}